\documentclass[final]{cvpr}

\usepackage{times}
\usepackage{epsfig}
\usepackage{graphicx}
\usepackage{amsmath}
\usepackage{amssymb}
\usepackage[sort,numbers]{natbib}
\usepackage{booktabs}
\usepackage{amsmath, bm, subfigure, epstopdf, url, pifont, overpic, cases}
\usepackage{latexsym, amssymb, bbding, multirow, makecell,  diagbox, enumitem, caption}
\usepackage{array, enumitem, soul, algorithm, algpseudocode}
\newcommand{\R}[1]{\textcolor[rgb]{1.00,0.00,0.00}{#1}}
\newcommand{\B}[1]{\textcolor[rgb]{0.00,0.00,1.00}{#1}}
\def\comp{\ensuremath\mathop{\scalebox{.6}{$\circ$}}}

\DeclareMathOperator*{\argmin}{arg\,min}

\usepackage{arydshln}

\usepackage[pagebackref=true,breaklinks=true,colorlinks,bookmarks=false]{hyperref}

\def\cvprPaperID{*} %
\def\confYear{*}

\begin{document}

\title{Flow-based Kernel Prior with Application to Blind Super-Resolution}
\author{Jingyun Liang$^{1}$ \qquad Kai Zhang$^{1,}$\thanks{Corresponding author.} \qquad Shuhang Gu$^{1,2}$ \qquad Luc Van Gool$^{1,3}$ \qquad Radu Timofte$^{1}$\\
$^{1}$ Computer Vision Lab, ETH Zurich, Switzerland\\
$^{2}$ The University of Sydney, Australia \quad $^{3}$ KU Leuven, Belgium\\
{\tt\small \{jinliang, kai.zhang, vangool, timofter\}@vision.ee.ethz.ch}\quad
{\tt\small shuhanggu@gmail.com}\\
\url{https://github.com/JingyunLiang/FKP}
}

\maketitle
\begin{abstract}
Kernel estimation is generally one of the key problems for blind image super-resolution (SR). Recently, Double-DIP proposes to model the kernel via a network architecture prior, while KernelGAN employs the deep linear network and several regularization losses to constrain the kernel space. However, they fail to fully exploit the general SR kernel assumption that anisotropic Gaussian kernels are sufficient for image SR. To address this issue, this paper proposes a normalizing flow-based kernel prior (FKP) for kernel modeling. By learning an invertible mapping between the anisotropic Gaussian kernel distribution and a tractable latent distribution, FKP can be easily used to replace the kernel modeling modules of Double-DIP and KernelGAN. Specifically, FKP optimizes the kernel in the latent space rather than the network parameter space, which allows it to generate reasonable kernel initialization, traverse the learned kernel manifold and improve the optimization stability. Extensive experiments on synthetic and real-world images demonstrate that the proposed FKP can significantly improve the kernel estimation accuracy with less parameters, runtime and memory usage, leading to state-of-the-art blind SR results.
\end{abstract}

\section{Introduction}
Image super-resolution (SR) is a fundamental low-level vision task whose goal is to recover the high-resolution (HR) image from the low-resolution (LR) input. With the development of convolutional neural networks (CNN), CNN-based methods~\cite{dong2014srcnn, kim2016vdsr, ledig2017srresnet, zhang2018rcan, wang2018esrgan, qiu2019ebrn, guo2020drn} have been gaining the popularity in solving image SR. However, most of existing works assume the blur kernel is fixed and known (\eg, bicubic downsampling kernel), which tends to result in a dramatic performance drop in real-world applications. Hence, blind image SR that aims to deal with unknown blur kernels is becoming an active research topic.

Compared to non-blind SR, blind SR generally needs to additionally estimate the blur kernel and thus is more ill-posed. A popular line of work tries to decompose blind SR into two sub-problems, \ie, kernel estimation and non-blind SR. As a preliminary step of non-blind SR, kernel estimation plays a crucial role. If the estimated kernel deviates from the ground-truth, the HR image reconstructed by the non-blind SR methods would deteriorate seriously~\cite{efrat2013accurate,yang2014single,gu2019sftmdikc}. In view of this, this paper focuses on the SR kernel estimation problem.

Recently, some kernel estimation methods, such as Double-DIP~\cite{gandelsman2019doubledip,ren2020selfdeblur} and KernelGAN~\cite{bell2019kernelgan}, have shown promising results. Specifically, with two deep image priors (DIPs)~\cite{ulyanov2018dip}, Double-DIP can be used to jointly optimize the HR image and blur kernel in the parameter space of untrained encoder-decoder networks by minimizing the LR image reconstruction error. Although DIP has shown to be effective for modeling natural images, whether it is effective to model blur kernel or not remains unclear. The main reason is that blur kernel usually has a small spatial size and has its own characteristics that differ from natural images. In~\cite{ren2020selfdeblur}, a fully-connected network (FCN) is used to model the kernel prior, which, however, lacks interpretability.
With a different framework to Double-DIP, KernelGAN designs an internal generative adversarial network (GAN) for the LR image on the basis of image patch recurrence property~\cite{glasner2009super,zontak2011internal,michaeli2013nonparametric}. It defines the kernel implicitly by a deep linear network, which is optimized by the GAN loss and five extra regularization losses such as sparsity loss. Obviously, these two methods do not make full use of the anisotropic Gaussian kernel prior which has been demonstrated to be effective enough for real image SR~\cite{efrat2013accurate, yang2014single, riegler2015cab, zhang2018srmd, kai2021bsrgan}.

In this paper, we propose a flow-based kernel prior (FKP) for kernel distribution modeling and incorporate it into existing blind SR models. Based on normalizing flow, FKP consists of several batch normalization layers, permutation layers and affine coupling layers, which allow the model to capture the kernel distribution by learning an invertible mapping between the kernel space and the latent space (\eg, high-dimensional Gaussian). FKP is optimized in an unsupervised way by minimizing the negative log-likelihood loss of the kernel. Once trained, it can be incorporated into existing blind SR models such as Double-DIP and KernelGAN for kernel estimation, in which FKP fixes its parameters and optimizes the latent variable in the network input space. Specifically, for Double-DIP, we jointly optimize DIP for HR image estimation and FKP for kernel estimation by minimizing the LR image reconstruction error. For KernelGAN, we blur the LR image with the kernel estimated by FKP rather than using a deep linear network, and then optimize it by adversarial training.

Using FKP as a kernel prior offers several advantages: 
1) \textit{Fewer parameters.}
FKP model only has 143K parameters, whereas Double-DIP and KernelGAN involve 641K and 151K for kernel modeling, respectively. 
2) \textit{More stable convergence.} On the one hand, unlike Double-DIP that uses random noise input and KernelGAN that uses random network parameters for kernel initialization, FKP can explicitly initialize a reasonable kernel since it is a bijective function. On the other hand, the kernel is implicitly constrained to be in the learned kernel manifold during model optimization.
3) \textit{Better kernel estimation.} With a learned kernel prior, the kernel estimation accuracy can be improved for several existing blind SR methods such as Double-DIP and KernelGAN.

The main contributions are summarized as follows:
\begin{itemize}
  \vspace{-0.2cm}
  \item[1)] We propose a kernel prior named FKP that is applicable for arbitrary blur kernel modeling. It learns a bijective mapping between the kernel and the latent variable. To the best of our knowledge, FKP is the first learning-based kernel prior.
  \vspace{-0.2cm}
  \item[2)] By fixing its parameters and optimizing the latent variable, FKP traverses the learned kernel manifold and searches for the kernel prediction, ensuring reasonable kernels for initialization and along optimization.
  \vspace{-0.2cm}
  \item[3)] With less parameters, runtime and memory usage, FKP improves the stability and accuracy of existing kernel estimation methods including Double-DIP and KernelGAN, leading to state-of-the-art blind SR performance.
\end{itemize}

\section{Related Work}
\vspace{-0.1cm}
\paragraph{Kernel Estimation.} 
Prior to the deep learning era, traditional kernel estimation methods typically utilize prior information of image patches or edges~\cite{begin2004blind, wang2005patch, he2009soft, michaeli2013nonparametric, shao2015simple}. In the deep learning era, Gandelsman~\etal~\cite{gandelsman2019doubledip} propose the Double-DIP based on the Deep Image Prior (DIP)~\cite{ulyanov2018dip}, which uses untrained encoder-decoder networks with skip connections as image priors for image dehazing, image deconvolution, transparency separation, \etc. Similarly, Ren~\etal~\cite{ren2020selfdeblur} propose a fully-connected network (FCN) as a kernel prior for image deconvolution. However, whether this idea works on blind SR kernel estimation or not is still an open problem, as blind SR is severely ill-posed due to downsampling. Different from above methods, Kligler~\etal~\cite{bell2019kernelgan} propose KernelGAN to estimate kernel based on the image patch recurrence property~\cite{glasner2009super, zontak2011internal, michaeli2013nonparametric}. They use a deep linear network as a generator to generate a re-downscaled image from the LR image, and a discriminator to ensure cross-scale patch similarity. The blur kernel is derived from the generator. Gu~\etal~\cite{gu2019sftmdikc} propose a predict-and-correct strategy to estimate kernel and HR image alternately. But it is highly dependent on training HR-LR image pairs and only estimates the feature of kernel.

\vspace{-0.3cm}
\paragraph{Normalizing Flow.}
Normalizing flows~\cite{dinh2014nice, dinh2016realnvp, kingma2016iaf, papamakarios2017maf, huang2018naf, kingma2018glow, jaini2019polyflow, lugmayr2020srflow} are invertible generative models that deform the complex data distribution to a simple and tractable distribution. Dinh~\etal~\cite{dinh2014nice} propose to stack non-linear additive coupling and other transformation layers as the flow model NICE. Inspired by NICE, Dinh~\etal~\cite{dinh2016realnvp} propose RealNVP, which upgrades additive coupling to affine coupling without loss of invertibility and achieves better performance. After that, Kingma~\etal~\cite{kingma2018glow} propose $1\times 1$ convolution to replace the fixed permutation layer in RealNVP and succeed to synthesize realistic-looking images. Normalizing flows have also been successfully applied in generating other types of data, such as audio data~\cite{kim2018flowavenet} and point cloud data~\cite{yang2019pointflow}.

\section{Flow-based Kernel Prior}
Generally, the classical degradation model of image SR~\cite{liu2013bayesian, elad1997restoration, farsiu2004advances} assumes the LR image $\mathbf{y}$ is obtained via a composition of blurring and downsampling from the HR image $\mathbf{x}$. 
Mathematically, it is formulated as
\begin{equation}
\mathbf{y} = (\mathbf{x} \otimes \mathbf{k}) \downarrow_s +  \mathbf{n},
\end{equation}
where $\mathbf{x}\otimes\mathbf{k}$ denotes the convolution between $\mathbf{x}$ and blur kernel $\mathbf{k}$, $\downarrow_s$ represents the downsampling operation with scale factor $s$, and $\mathbf{n}$ is the noise. Particularly, blind SR~\cite{bell2019kernelgan,gu2019sftmdikc,he2009soft,ulyanov2018dip} aims to estimate the HR image and blur kernel simultaneously. According to the \emph{Maximum A Posteriori} (MAP) framework, it can be solved as 
\begin{equation}
\mathbf{x}^*, \mathbf{k}^* = \argmin_{\mathbf{x},  \mathbf{k}} \|\mathbf{y}-(\mathbf{x} \otimes \mathbf{k}) \downarrow_s\|^2  +\lambda\Phi(\mathbf{x}) +\gamma\Omega(\mathbf{k}),
\label{eq:map_original}
\end{equation}%
where $\|\mathbf{y}-(\mathbf{x} \otimes \mathbf{k}) \downarrow_s\|^2$ is the data fidelity term, $\Phi(\mathbf{x})$ denotes the image prior, $\Omega(\mathbf{k})$ represents the kernel prior, $\lambda$ and $\gamma$ are trade-off parameters. It has been well-studied that poor kernel estimation would cause a severe performance drop for HR image estimation~\cite{efrat2013accurate, gu2019sftmdikc, zhang2018srmd}. However, while various image priors have been proposed to describe natural image statistics~\cite{chan1998tv, candes2008enhancing, dabov2007image, he2010darkchannel, ulyanov2018dip, pan2020dgp}, little attention has been paid on designing the kernel prior.

\begin{figure}[!tbp]
\captionsetup{font=small}%
\begin{center}
\begin{overpic}[width=7.5cm]{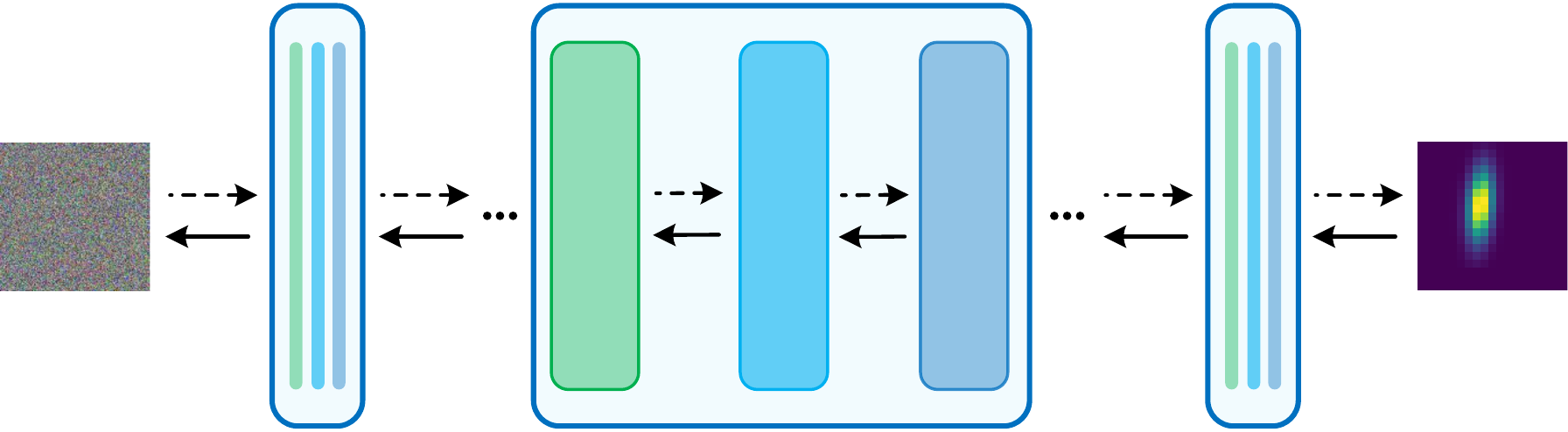}
\put(3,5){\color{black}{\scriptsize $\mathbf{z}_\mathbf{k}$}}
\put(94,5){\color{black}{\scriptsize $\mathbf{k}$}}
\put(41,-3){\color{black}{ \fontsize{7}{5}\selectfont {Flow Block}}}
\put(59,5.1){\color{black}{ \fontsize{7}{5}\selectfont \rotatebox{90}{ {\makecell{Batch Norm.}}}}}
\put(47.5,5.3){\color{black}{ \fontsize{7}{5}\selectfont \rotatebox{90}{ {\makecell{Permutation}}}}}
\put(35.5,5.0){\color{black}{ \fontsize{7}{5}\selectfont \rotatebox{90}{ {\makecell{Affine Trans.}}}}}

\end{overpic}
\end{center}\vspace{-0.1cm}
\caption{The schematic illustration of the flow-based kernel prior (FKP) network. FKP learns an invertible mapping between the kernel $\mathbf{k}$ and the latent variable $\mathbf{z}_\mathbf{k}$ by several flow blocks, each of which is a succession of batch normalization, permutation and affine transformation layers.}
\label{fig:fkp}
\vspace{-0.1cm}
\end{figure}
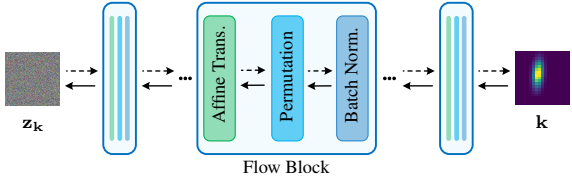

In view of this, we aim to learn a kernel prior based on normalizing flow in this paper. Formally, let $\mathbf{k}\in K$ denotes the kernel variable and $\mathbf{z}_\mathbf{k}\in Z$ denotes the corresponding latent variable. $\mathbf{k}$ and $\mathbf{z}_\mathbf{k}$ obey probability distributions $p_K$ and $p_Z$, respectively. We define a bijection $f_{\bm{\theta}}: K \rightarrow Z$ with parameter $\bm{\theta}$. For kernel $\mathbf{k}$, it can be encoded as a latent variable $\mathbf{z}_\mathbf{k}=f_{\bm{\theta}}(\mathbf{k})$ in the latent space. Inversely, $\mathbf{k}$ could be exactly reconstructed by the inverse mapping: $\mathbf{k}=f_{\bm{\theta}}^{-1}(\mathbf{z}_\mathbf{k})$. According to the \emph{change of variable} formula~\cite{dinh2014nice}, the probability of $\mathbf{k}$ is computed as
\begin{equation}
p_K (\mathbf{k}) = p_Z (f_{\bm{\theta}}(\mathbf{k})) 
\left\lvert 
\det(\frac{\partial f_{\bm{\theta}}(\mathbf{k})}{\partial \mathbf{k}}) \right\rvert,
\end{equation}%
where $\frac{\partial f_{\bm{\theta}}(\mathbf{k})}{\partial \mathbf{k}}$ is the Jacobian of $f_{\bm{\theta}}$ at $\mathbf{k}$. Generally, $p_Z$ is a simple tractable distribution, such as multivariate Gaussian distribution. $f_{\bm{\theta}}$ is often composed of a sequence of invertible and tractable transformations: $f_{\bm{\theta}} = f_{\bm{\theta}}^1 \comp f_{\bm{\theta}}^2 \comp \cdots \comp f_{\bm{\theta}}^N$, and we have $\mathbf{h}^{n}=f_{\bm{\theta}}^n(\mathbf{h}^{n-1})$ for $n\in\{1,...,N\}$. The input and output $\mathbf{h}^{0}$ and $\mathbf{h}^{N}$ of $f_{\bm{\theta}}$ are $\mathbf{k}$ and $\mathbf{z}_\mathbf{k}$, respectively. Under maximum likelihood estimation, $\bm{\theta}$ can be optimized by minimizing the negative log-likelihood (NLL) loss
\begin{equation}
\begin{aligned}
\mathcal{L} (\mathbf{k}; \bm\theta) 
&= -\log p_Z (f_{\bm{\theta}}(\mathbf{k})) 
-\sum_{n=1}^{N} \log
\left\lvert 
\det(\frac{\partial f_{\bm{\theta}}^n (\mathbf{h}^{n-1})}{\partial \mathbf{h}^{n-1}}) 
\right\rvert.
\end{aligned}
\label{eq:nll}
\end{equation}%

More specifically, we build FKP by stacking invertible flow layers. As shown in Fig.~\ref{fig:fkp}, it consists of several flow blocks, and each block includes three consecutive layers: batch normalization layer, permutation layer and affine transformation layer~\cite{dinh2016realnvp}. For affine transformation layer, we use small fully-connected neural networks (FCN) for scaling and shifting, in which each FCN stacks fully-connected layers and \emph{tanh} activation layers alternately.

\begin{figure}[!t]
\captionsetup{font=small}%
\scriptsize %
\centering
\hspace{-0.5cm}
\begin{tabular}{m{0.65cm}<{\centering}m{0.65cm}<{\centering}m{0.65cm}<{\centering}m{0.65cm}<{\centering}m{0.65cm}<{\centering}m{0.65cm}<{\centering}m{0.65cm}<{\centering}}
		
		\includegraphics[width=0.06\textwidth]{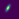}~~
		&\includegraphics[width=0.06\textwidth]{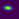}~~
		&\includegraphics[width=0.06\textwidth]{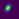}~~
		&\includegraphics[width=0.06\textwidth]{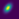}~~
		&\includegraphics[width=0.06\textwidth]{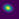}~~
		&\includegraphics[width=0.06\textwidth]{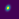}~~
		&        \includegraphics[width=0.06\textwidth]{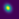}~\\ 
		
	\end{tabular}
    \vspace{-0.1cm}
	\caption{Diverse kernel samples generated by randomly sampling from the latent space of the flow-based kernel prior (FKP). Here, FKP is trained on anisotropic Gaussian kernels with scale factor 4.}
	\label{fig:kernel_random}
	\vspace{-0.1cm}
\end{figure}

FKP is trained by the NLL loss given training kernel samples. When it is plugged into existing kernel estimation models as a kernel prior, we first randomly sample a latent variable $\mathbf{z}_\mathbf{k}$, which corresponds to a random kernel as shown in Fig.~\ref{fig:kernel_random}. Then, we fix the model parameters and update $\mathbf{z}_\mathbf{k}$ by gradient back-propagation under the guidance of kernel estimation loss. Instead of starting with random initialization and slowly updating the kernel, FKP moves along the learned kernel manifold and generates reliable kernels $f_{\bm{\theta}}^{-1}(\mathbf{z}_\mathbf{k})$ during the update of $\mathbf{z}_\mathbf{k}$. In addition, when $\mathbf{z}_\mathbf{k}$ follows multivariate Gaussian distribution, most of the mass of distribution is near the surface of a sphere of radius $\sqrt{D}$~\cite{vershynin2018random, menon2020pulse}, where $D$ is the dimension of $\mathbf{z}_\mathbf{k}$. Therefore we optimize $\mathbf{z}_\mathbf{k}$ on the sphere surface by restricting its Euclidean norm as $\left\|\mathbf{z}_\mathbf{k}\right\|_2 =\sqrt{D}$ after every update, avoiding optimizing in the entire latent space.

\section{Incorporating FKP to Double-DIP}
\subsection{Original Double-DIP}
\label{sec:original_doubledip}
DIP~\cite{ulyanov2018dip} is an image prior in the form of a randomly initialized encoder-decoder network $\mathcal{G}$, in which the network structure captures low-level image statistics. By optimizing network parameter $\bm\theta_\mathcal{G}$, natural-looking image $\mathbf{x} = \mathcal{G}(\mathbf{z}_\mathbf{x}; \bm\theta_\mathcal{G})$ is reconstructed from the fixed random noise input $\mathbf{z}_\mathbf{x}$. To model different image components, Double-DIP~\cite{gandelsman2019doubledip} couples two DIPs for image decomposition tasks such as image segmentation. This framework is also exploited in image deconvolution by replacing one DIP with a fully-connected network (FCN)~\cite{ren2020selfdeblur}. If Double-DIP and its variants are used for blind SR, they can be formulated as
\begin{numcases}{\hspace{-0.5cm}}\hspace{-0.1cm}
\bm\theta_\mathcal{G}^*,\bm\theta_\mathcal{K^\prime}^* = \argmin\limits_{\bm\theta_\mathcal{G},\bm\theta_\mathcal{K^\prime}}  \|\mathbf{y}-(\mathcal{G}(\mathbf{z}_{\mathbf{x}};\bm\theta_\mathcal{G}) \otimes \mathcal{K^\prime(\mathbf{z}_\mathbf{k};\bm\theta_\mathcal{K^\prime}}))\downarrow_{s}\|^2\nonumber\hspace{-0.5cm}\\
\hspace{-0.1cm}\mathbf{x}^* = \mathcal{G}(\mathbf{z}_\mathbf{x};\bm\theta_\mathcal{G}^*), \quad \mathbf{k}^* = \mathcal{K^\prime(\mathbf{z}_\mathbf{k};\bm\theta_\mathcal{K^\prime}^*})\hspace{-0.5cm}
\end{numcases}
where $\mathcal{K^\prime(\mathbf{z}_\mathbf{k};\bm\theta_\mathcal{K^\prime}})$ is a kernel prior based on untrained neural networks such as DIP and FCN. In training, random noise inputs $\mathbf{z}_\mathbf{x}$ and $\mathbf{z}_\mathbf{k}$ are fixed, while randomly initialized network parameter $\bm\theta_\mathcal{G}$ and $\bm\theta_\mathcal{K^\prime}$ are optimized to minimize the LR image reconstruction error. 

However, applying above Double-DIP framework to blind SR kernel estimation is non-trivial. On the one hand, unlike images, kernels are spatially small and have no natural image properties such as self-similarity. Designing a DIP-like network based on convolution layers or fully-connected layers and then using the network architecture as a kernel prior may not be a good choice. On the other hand, due to the downsampling operation, blind SR is extremely ill-posed compared with other image restoration tasks. The knowledge incorporated by untrained networks may not be sufficient for estimating the HR image and kernel simultaneously. As will be shown in experiments, untrained neural networks fail to generate reasonable kernel estimations.

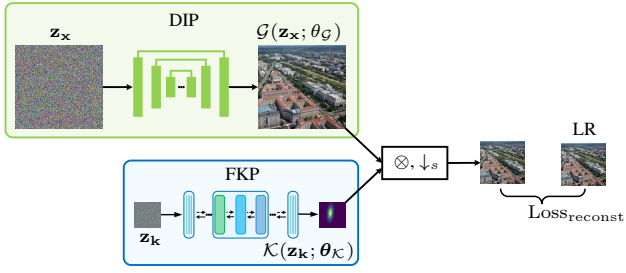
\begin{figure}[!tbp]
\captionsetup{font=small}%
\begin{center}
\begin{overpic}[width=8cm]{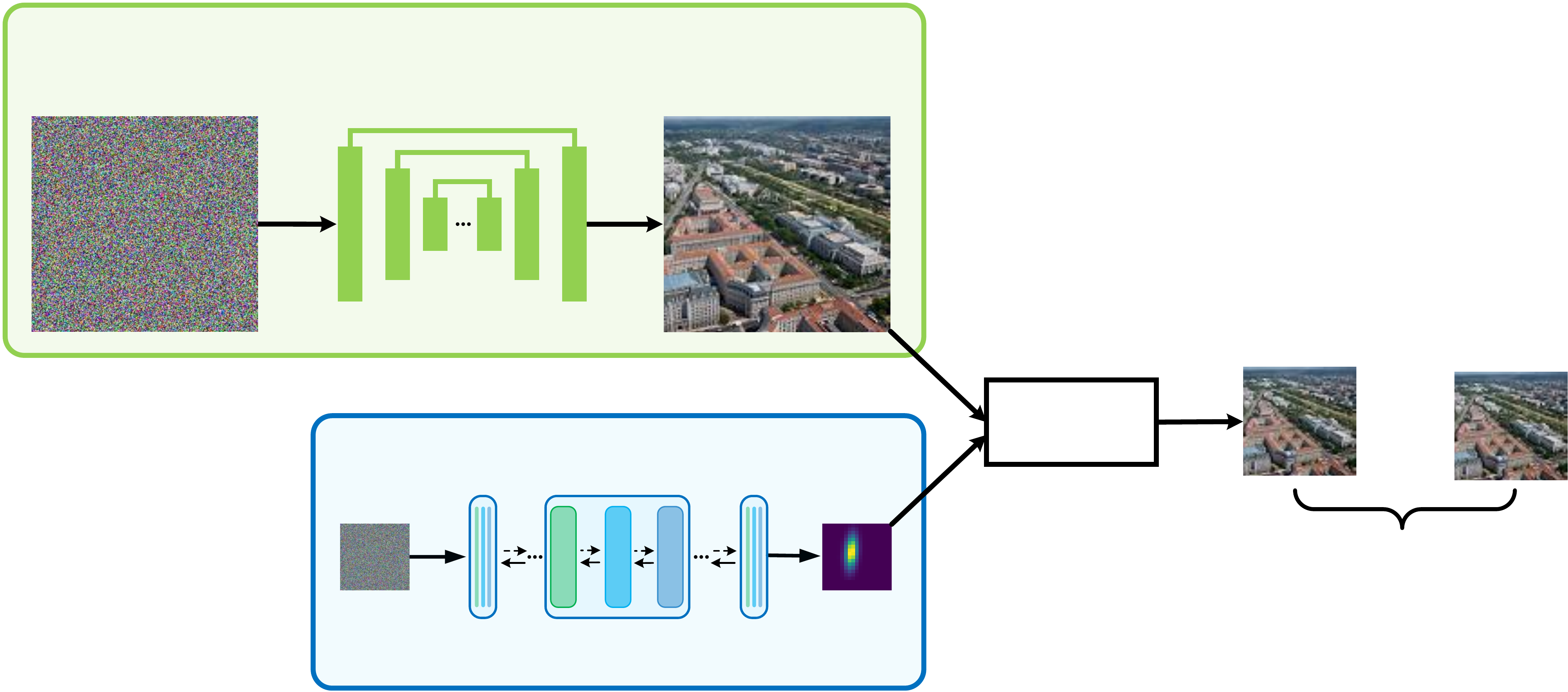}
\put(7.7,38.1){\color{black}{\scriptsize $\mathbf{z}_\mathbf{x}$}}
\put(27.5,40){\color{black}{\fontsize{7}{5em}\selectfont DIP}}
\put(42,38.2){\color{black}{\scriptsize $\mathcal{G}(\mathbf{z}_\mathbf{x};\theta_\mathcal{G})$}}

\put(63.7,16.6){\color{black}{ \scriptsize $\otimes$, $\downarrow_s$}}
\put(93.5,21.8){\color{black}{ \fontsize{7}{5em}\selectfont {LR}}}
\put(86,8){\color{black}{ \fontsize{7}{5em}\selectfont  {$\rm{Loss_{reconst}}$}}}

\put(21.5,4){\color{black}{ \scriptsize $\mathbf{z}_\mathbf{k}$}}
\put(35.8,14.3){\color{black}{ \fontsize{7}{5em}\selectfont {FKP}}}
\put(42,1.8){\color{black}{ \scriptsize $\mathcal{K(\mathbf{z}_\mathbf{k};\bm\theta_\mathcal{K}})$}}
\end{overpic}
\end{center}\vspace{-0.3cm}
\caption{The schematic illustration of DIP-FKP. DIP estimates SR image $\mathcal{G}(\mathbf{z}_\mathbf{x};\theta_\mathcal{G})$ from $\mathbf{z}_\mathbf{x}$, while FKP estimates kernel $\mathcal{G}(\mathbf{z}_\mathbf{x};\theta_\mathcal{G})$ from $\mathbf{z}_\mathbf{k}$. $\otimes$ and $\downarrow_s$ denote blur and downsampling with scale factor $s$, respectively. The model is optimized by minimizing the LR image reconstruction error.}
\label{fig:dip_fkp}
\vspace{-0.1cm}
\end{figure}

\subsection{Proposed DIP-FKP}
Instead of using a DIP-like untrained network as the kernel prior, we propose to incorporate FKP into the Double-DIP framework, which we refer to as DIP-FKP. The HR image and kernel are jointly estimated as
\begin{numcases}{\hspace{-0.2cm}}\hspace{-0.1cm}
\bm\theta_\mathcal{G}^*, \mathbf{z}_\mathbf{k}^* = \argmin\limits_{\bm\theta_\mathcal{G},\mathbf{z}_\mathbf{k}}  \|\mathbf{y}-(\mathcal{G}(\mathbf{z}_{\mathbf{x}};\bm\theta_\mathcal{G}) \otimes \mathcal{K(\mathbf{z}_\mathbf{k};\bm\theta_\mathcal{K}}))\downarrow_{s}\|^2 \nonumber\\
\hspace{-0.1cm}\mathbf{x}^* = \mathcal{G}(\mathbf{z}_\mathbf{x};\bm\theta_\mathcal{G}^*), \quad \mathbf{k}^* = \mathcal{K(\mathbf{z}_\mathbf{k}^*;\bm\theta_\mathcal{K}})
\end{numcases}
where $\mathcal{K(\mathbf{z}_\mathbf{k};\bm\theta_\mathcal{K}})$ is the incorporated FKP. The corresponding schematic illustration is shown in Fig.~\ref{fig:dip_fkp}. 

In DIP-FKP, we optimize the kernel latent variable $\mathbf{z}_\mathbf{k}$, rather than network parameter $\bm\theta_\mathcal{K}$, which is fixed as it has modeled the kernel prior. More specifically, in forward propagation, FKP generates a kernel prediction $\mathcal{K(\mathbf{z}_\mathbf{k};\bm\theta_\mathcal{K}})$ to blur the SR image $ \mathcal{G}(\mathbf{z}_\mathbf{x}; \bm\theta_\mathcal{G})$ produced by DIP, in order to obtain the LR image prediction. The mean squared error between the resulting LR prediction and LR image are used as the loss function. In back propagation, gradients are back-propagated from the loss function to kernel prediction, and then to the latent variable $\mathbf{z}_\mathbf{k}$.

With FKP, DIP-FKP embeds kernel prior into the network effectively by moving the kernel prediction along the learned kernel manifold, which enables accurate and stable kernel estimation. Therefore, without massive training data and long training time, DIP-FKP can estimate SR image and blur kernel simultaneously during testing phase. It is noteworthy that while DIP-FKP is able to estimate kernels accurately, it has limited performance on SR image reconstruction as it is self-supervised. For this reason, we use non-blind model USRNet~\cite{zhang2020usrnet} to generate the final SR result based on the kernel estimation.

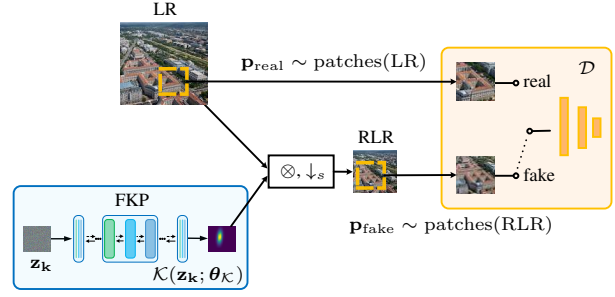
\begin{figure}[!tbp]
\captionsetup{font=small}%
\begin{center}
\begin{overpic}[width=8cm]{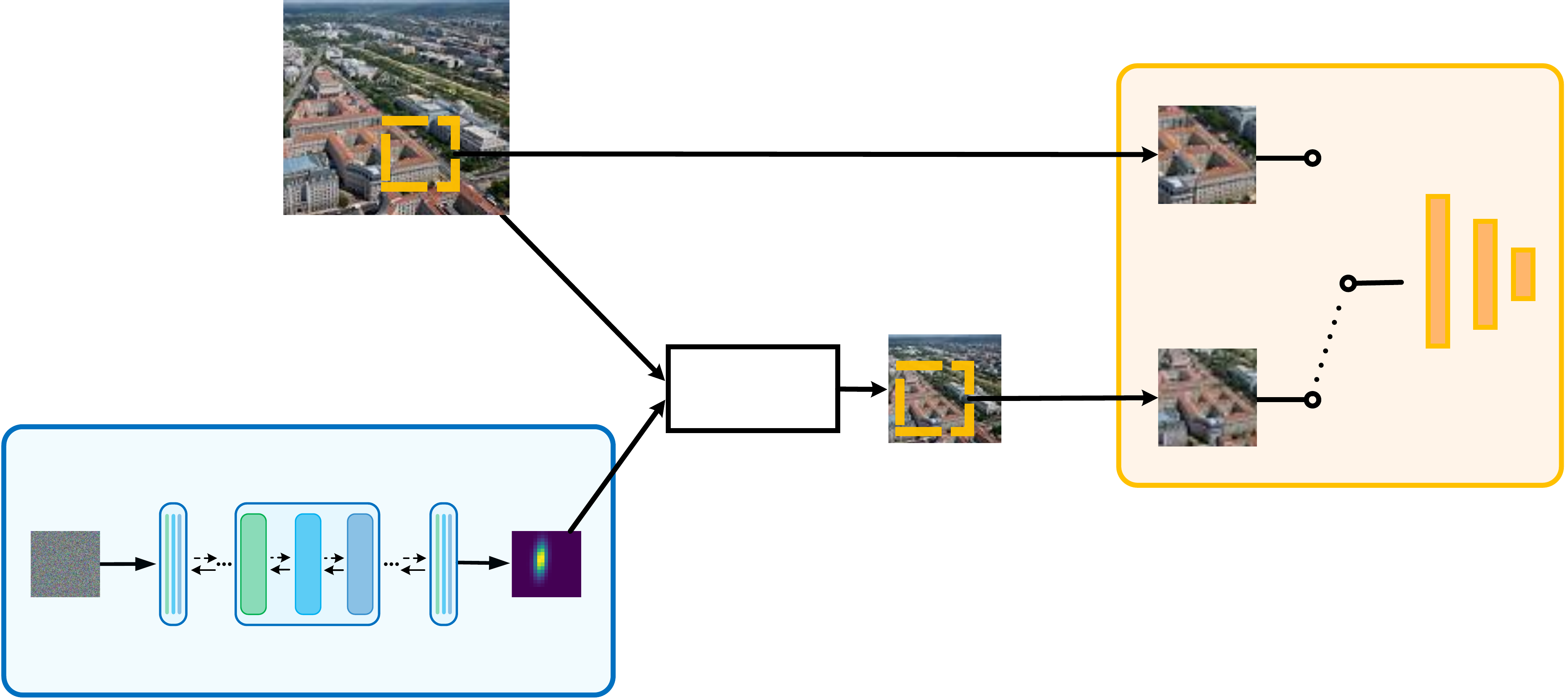}
\put(23.5,45.8){\color{black}{\fontsize{7}{5em}\selectfont {LR}}}
\put(38,37.2){\color{black}{\fontsize{7}{5em}\selectfont $\mathbf{p}_{\rm real}\sim {\rm  {patches}}({\rm  {LR}})$}}
\put(85,33.8){\color{black}{\scriptsize {\rm real}}}
\put(85,18.2){\color{black}{\scriptsize {\rm fake}}}

\put(43.4,19){\color{black}{ \scriptsize $\otimes$, $\downarrow_s$}}
\put(93,36){\color{black}{ \fontsize{7}{5em}\selectfont {$\mathcal{D}$}}}

\put(1.9,3.7){\color{black}{ \scriptsize $\mathbf{z}_\mathbf{k}$}}
\put(16,14.1){\color{black}{ \fontsize{7}{5em}\selectfont {FKP}}}
\put(22.2,1.8){\color{black}{ \scriptsize $\mathcal{K(\mathbf{z}_\mathbf{k};\bm\theta_\mathcal{K}})$}}
\put(57.5,24.5){\color{black}{\fontsize{7}{5em}\selectfont {RLR}}}
\put(56,10){\color{black}{\fontsize{7}{5em}\selectfont $\mathbf{p}_{\rm{fake}}\sim {\rm  {patches}}({\rm  {RLR}})$}}
\end{overpic}
\end{center}\vspace{-0.3cm}
\caption{The schematic illustration of KernelGAN-FKP. Kernel $\mathcal{K(\mathbf{z}_\mathbf{k};\bm\theta_\mathcal{K}})$ is generated by FKP and is then used to generate the re-downscaled LR (RLR) image from the LR image. $\otimes$ and $\downarrow_s$ denote blur and downsampling with scale factor $s$, respectively. The model is optimized similar to GAN: FKP tries to fool the discriminator $\mathcal{D}$ to believe that image patch $\mathbf{p}_{\rm real}$ extracted from the LR image and patch $\mathbf{p}_{\rm fake}$ extracted from the RLR image share the same distribution.}
\label{fig:kernelgan_fkp}
\end{figure}

\section{Incorporating FKP to KernelGAN}
\subsection{Original KernelGAN}
In a single image, small image patches tend to recur across different scales~\cite{glasner2009super, zontak2011internal}. It is further observed in~\cite{michaeli2013nonparametric} that when the LR image is degraded with a blur kernel to obtain a re-downscaled low-resolution (RLR) image, the kernel between the HR and LR image maximizes the internal patch distribution similarity between the LR and RLR image. According to this observation, KernelGAN~\cite{bell2019kernelgan} trains an internal generative adversarial network (GAN) on a single LR image in order to estimate the blur kernel. It consists of a deep linear generator $\mathcal{G}$ that downscales the LR image by several learnable convolution layers and a discriminator $\mathcal{D}$ that distinguishes between patch distributions of the LR and RLR image. The blur kernel is derived from $\mathcal{G}$ in every iteration. Formally, KernelGAN is optimized as 
\begin{numcases}{\hspace{-0.3cm}}\hspace{-0.1cm}
\bm\theta_\mathcal{G}^*, \bm\theta_\mathcal{D}^*=
\argmin\limits_{\bm\theta_\mathcal{G}}\max\limits_{\bm\theta_\mathcal{D}}
\left\{ (\mathcal{D}(\mathbf{p})-1)^2 
+(\mathcal{D}(\mathcal{G}(\mathbf{p})))^2+\mathcal{R}\right \}\nonumber\hspace{-0.5cm}\\
\hspace{-0.1cm}\mathbf{k}^* \leftarrow \bm\theta_\mathcal{G}^*\hspace{-0.5cm}
\end{numcases}
where $\mathbf{p}$ is a patch randomly extracted from the LR image $\mathbf{y}$, \ie, $\mathbf{p}\sim {\rm patches}(\mathbf{y})$, and $\mathcal{R}$ represents extra regularization on the kernel $\mathbf{k}$. More specifically, $\mathcal{R}$ includes the mean squared error between $\mathbf{k}$ and the bicubic kernel, as well as other regularization constraints on kernel pixel sum, boundary, sparsity and centrality. However, the performance of KernelGAN is unstable. It also suffers from the burden of hyper-parameter selection due to multiple regularization terms.

\subsection{Proposed KernelGAN-FKP}
The instability of KernelGAN may come from the weak patch distribution of some images, \ie, there are multiple kernels that could generate the RLR image with similar patch distribution to the LR image. In this case, the discriminator cannot distinguish between different patch distributions, thus leading to wrong kernel prediction. To alleviate the problem, we propose to incorporate the proposed FKP into KernelGAN in order to constrain the optimization space. We refer to this method as KernelGAN-FKP which can be formulated as
\begin{equation}
\begin{cases}
\mathbf{z}_\mathbf{k}^*, \bm\theta_\mathcal{D}^*=
\argmin\limits_{\mathbf{z}_\mathbf{k}}\max\limits_{\bm\theta_\mathcal{D}}
\left\{ (\mathcal{D}(\mathbf{p})-1)^2 \right .\\
\left. \qquad\qquad\qquad~~+(\mathcal{D}((\mathbf{p}\otimes \mathcal{K(\mathbf{z}_\mathbf{k};\bm\theta_\mathcal{K}}))\downarrow_s))^2 \right \}\\
\mathbf{k}^* = \mathcal{K(\mathbf{z}_\mathbf{k}^*;\bm\theta_\mathcal{K}})
\label{eq:kernelganfkp}
\end{cases}
\end{equation}

As illustrated schematically in Fig.~\ref{fig:kernelgan_fkp}, KernelGAN-FKP directly generates the kernel from the latent variable $\mathbf{z}_\mathbf{k}$ and uses it to degrade the LR image instead of using a deep linear network. In optimization, $\mathbf{z}_\mathbf{k}$ is optimized to fool the discriminator, which equals to traversing in the kernel space to find a kernel that the discriminator cannot distinguish. This constrains the optimization space of the generator and ensures kernel generation quality, which allows more stable convergence than the original KernelGAN, even without extra regularization terms. Similar to DIP-FKP, we adopt USRNet~\cite{zhang2020usrnet} for non-blind SR after kernel prediction.

\begin{table*}[thbp]
\captionsetup{font=small}
\scriptsize
\center
\begin{center}
\caption[Caption for LOF]{Average PSNR/SSIM of different methods on various datasets. Note that due to GPU memory constraints, we crop $960\times 960$ center image patches for DIV2K in kernel estimation. The best and second best results are highlighted in \R{red} and \B{blue} colors, respectively.}
\label{tab:dipfkp_psnr}
\begin{tabular}{|l|c|c|c|c|c|c|}
\hline
Method & Scale &  Set5~\cite{Set5} &  Set14~\cite{Set14} &  BSD100~\cite{BSD100} &  Urban100~\cite{Urban100} &  DIV2K~\cite{DIV2K}
\\
\hline
\hline %
Bicubic Interpolation & $\times$2 
& 26.58/0.8010 & 24.85/0.6939 & 25.19/0.6633 & 22.35/0.6503 & 26.97/0.7665  \\
RCAN~\cite{zhang2018rcan} & $\times$2 
& 26.80/0.8004 & 24.83/0.6945 & 25.21/0.6619 & 22.30/0.6499 & 26.99/0.7666  \\
DIP~\cite{ulyanov2018dip} & $\times$2 
& 26.82/0.7518 & 25.40/0.6868 & 24.71/0.6508 & 23.29/0.6749 & -  \\
Double-DIP~\cite{gandelsman2019doubledip} & $\times$2 
& 24.71/0.6423 & 22.21/0.5626 & 23.31/0/5681 & 21.03/0.5701 & -  \\
DIP-FKP (ours) & $\times$2 
& \B{30.16/0.8637} & \B{27.06/0.7421} & \B{26.72/0.7089} & \B{24.33/0.7069} & {-}  \\
DIP-FKP + USRNet~\cite{zhang2020usrnet} (ours) & $\times$2 
& \R{32.34/0.9308} & \R{28.18/0.8088} & \R{28.61/0.8206} & \R{26.46/0.8203} & \R{30.13/0.8686}  \\
\hdashline
GT + USRNet~\cite{zhang2020usrnet} (upper bound)  & $\times$2  & 36.37/0.9508 & 32.56/0.8945 & 31.34/0.8772 & 29.97/0.8954 & 34.59/0.9268 \\
\hline
\hline %
Bicubic Interpolation & $\times$3 
& 23.38/0.6836 & 22.47/0.5884 & 23.17/0.5625 & 20.37/0.5378 & 24.50/0.6806  \\
RCAN~\cite{zhang2018rcan} & $\times$3 
& 23.56/0.6802 & 22.31/0.5801 &  23.04/0.5506 & 20.14/0.5247 & 24.32/0.6712  \\
DIP~\cite{ulyanov2018dip} & $\times$3 
& 28.14/0.7687 & 25.19/0.6581 & 25.25/0.6408 & 23.22/0.6512 & -  \\
Double-DIP~\cite{gandelsman2019doubledip} & $\times$3 
& 23.21/0.6535 & 20.20/0.5071 & 20.38/0.4499 & 19.61/0.4993 & -  \\
DIP-FKP (ours) & $\times$3 
& \B{28.82/0.8202} & \B{26.27/0.6922} & \B{25.96/0.6660} &  \B{23.47/0.6588} & {-}  \\
DIP-FKP + USRNet~\cite{zhang2020usrnet} (ours) & $\times$3 
& \R{30.78/0.8840} & \R{27.76/0.7750} & \R{27.29/0.7484} & \R{24.84/0.7510} & \R{29.03/0.8354}  \\
\hdashline
GT + USRNet~\cite{zhang2020usrnet} (upper bound)  & $\times$3  & 33.95/0.9199 & 29.91/0.8283 & 28.82/0.7935 & 27.22/0.8274 & 31.79/0.8754 \\
\hline
\hline %
Bicubic Interpolation & $\times$4 
& 21.70/0.6198 & 20.86/0.5181 & 21.95/0.5097 & 19.13/0.4729 & 23.01/0.6282 \\
RCAN~\cite{zhang2018rcan} & $\times$4 
& 21.86/0.6174 & 20.37/0.4940 & 21.71/0.4935 & 18.60/0.4465 & 22.69/0.6128  \\
DIP~\cite{ulyanov2018dip} & $\times$4 
& 27.34/0.7465 & 25.03/0.6371 & 24.92/0.6030 & 22.55/0.6128 & -  \\
Double-DIP~\cite{gandelsman2019doubledip} & $\times$4 
& 20.99/0.5578 & 18.31/0.4426 & 18.57/0.3815 & 18.15/0.4491 & -  \\
DIP-FKP (ours) & $\times$4 
& \B{27.77/0.7914} & \B{25.65/0.6764} & \B{25.15/0.6354} & \B{22.89/0.6327} & {-}  \\
DIP-FKP + USRNet~\cite{zhang2020usrnet} (ours) & $\times$4 
& \R{29.29/0.8508} & \R{26.70/0.7383} & \R{25.97/0.6902} & \R{23.89/0.7078} & \R{27.44/0.7859}  \\
\hdashline
GT + USRNet~\cite{zhang2020usrnet} (upper bound)  & $\times$4  & 31.91/0.8894 & 28.30/0.7742 & 27.33/0.7277 & 25.47/0.7635 & 29.99/0.8272 \\
\hline             
\end{tabular}
\end{center}
\end{table*}

\section{Experiments}
\subsection{Experimental Setup}
\vspace{-0.1cm}
\paragraph{Data Preparation.}
Since the blur kernels of real LR images are usually unimodal~\cite{michaeli2013nonparametric, shao2015simple} and can typically be modeled by a Gaussian~\cite{riegler2015cab}, most existing blind SR works~\cite{he2009soft, bell2019kernelgan,gu2019sftmdikc,riegler2015cab,shocher2018zssr,soh2020mzsr,zhang2018srmd,xu2020udvd, kai2021bsrgan} assume the SR kernel is isotropic or anisotropic Gaussian kernel. Following this widely-adopted assumption, we conduct experiments on anisotropic Gaussian kernels, even though FKP is trained in an unsupervised way and can be used to model arbitrary kernel estimation given kernel samples. For scale factor $s\in \{2,3,4\}$, the kernel sizes and width ranges are set to $(4s+3)\times(4s+3)$ and $[0.175s,2.5s]$, respectively. The rotation angle range is $[0,\pi]$ for all $s$. We blur and downsample images with random kernels to generate testing sets based on Set5~\cite{Set5}, Set14~\cite{Set14}, BSD100~\cite{BSD100}, Urban100~\cite{Urban100} and the validation set of DIV2K~\cite{DIV2K}. Following KernelGAN~\cite{bell2019kernelgan} and USRNet~\cite{zhang2020usrnet}, the blur kernel is shifted and the upper-left pixels are kept in downsampling to avoid sub-pixel misalignments. For evaluation, we compare kernels by PSNR, and compare SR images by PSNR and SSIM~\cite{SSIM} on Y channel in the YCbCr space.

\vspace{-0.3cm}
\paragraph{FKP.}
Since FKP is a bijection, the latent variable dimension, the FCN input and output dimensions are the same as the kernel size. To capture kernel variety, the dimension of FCN hidden layers is set to $5(s+1)$ for scale factor $s$. The total number of flow blocks and FCN depth are set to 5 and 3, respectively. We randomly generate anisotropic Gaussian kernels as training data and use the Adam optimizer~\cite{kingma2014adam} to optimize the model for 50,000 iterations. The batch size and learning rate are set to 100 and 1e-4, respectively. Due to page limit, ablation study on FKP is provided in the supplementary.

\vspace{-0.3cm}
\paragraph{DIP-FKP.}
The DIP architecture is same as in \cite{ulyanov2018dip}. The model is optimized by the Adam optimizer for 1,000 iterations with $\beta_1=0.9$ and $\beta_2=0.999$. The learning rates for FKP and DIP are set to 0.1 and 0.005, respectively.

\vspace{-0.3cm}
\paragraph{KernelGAN-FKP.}
Small patches of size $64 \times 64$ are randomly cropped from the LR image for GAN training. We use the WGAN-GP loss~\cite{gulrajani2017improved} and set the gradient penalty to 0.1 to increase training stability. The learning rate is 5e-4 and the batch size is 64. For scale factors 2 and 4, we train the model for 1,000 and 4,000 iterations, respectively, by the Adam optimizer with $\beta_1=0.5$ and $\beta_2=0.999$.

\begin{figure*}[htbp]
\captionsetup{font=small}
\scriptsize
\begin{tabular}{c@{\extracolsep{0em}}@{\extracolsep{0em}}c@{\extracolsep{0em}}c@{\extracolsep{0em}}c@{\extracolsep{0em}}c@{\extracolsep{0em}}c@{\extracolsep{0em}}@{\extracolsep{0em}}c}
		\includegraphics[width=0.137\textwidth]{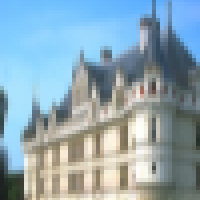}~
		&\includegraphics[width=0.137\textwidth]{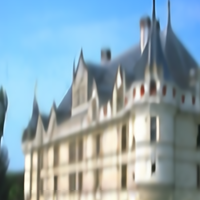}~
		&\includegraphics[width=0.137\textwidth]{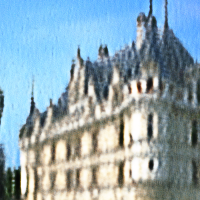}~
        &\includegraphics[width=0.137\textwidth]{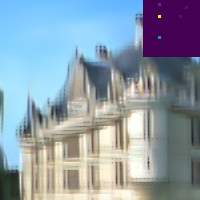}~
        &\includegraphics[width=0.137\textwidth]{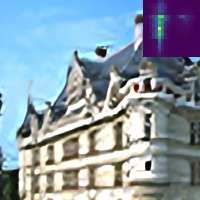}~
		&\includegraphics[width=0.137\textwidth]{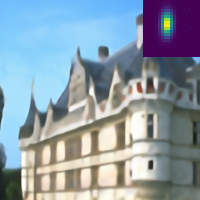}~
		&\includegraphics[width=0.137\textwidth]{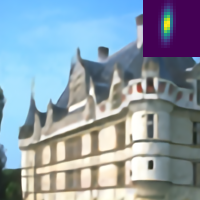}~\\
	    
        \includegraphics[width=0.137\textwidth]{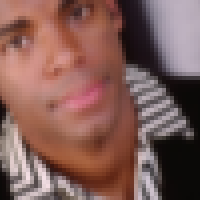}~
		&\includegraphics[width=0.137\textwidth]{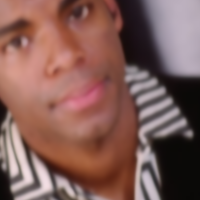}~
		&\includegraphics[width=0.137\textwidth]{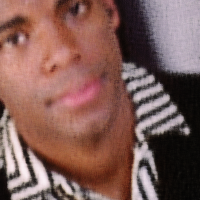}~
        &\includegraphics[width=0.137\textwidth]{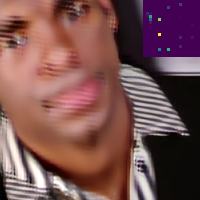}~
        &\includegraphics[width=0.137\textwidth]{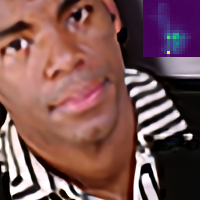}~
		&\includegraphics[width=0.137\textwidth]{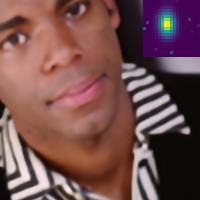}~
	    &\includegraphics[width=0.137\textwidth]{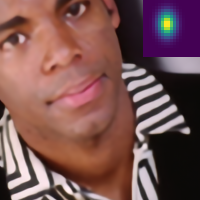}~\\

		\includegraphics[width=0.137\textwidth]{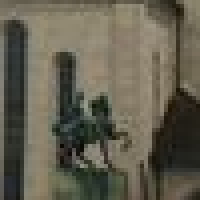}~
		&\includegraphics[width=0.137\textwidth]{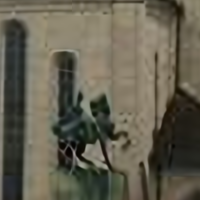}~
		&\includegraphics[width=0.137\textwidth]{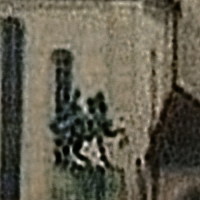}~
        &\includegraphics[width=0.137\textwidth]{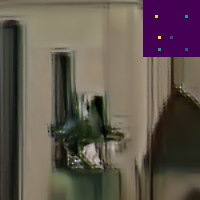}~
        &\includegraphics[width=0.137\textwidth]{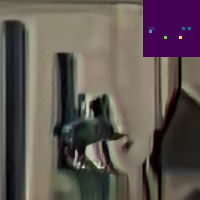}~
		&\includegraphics[width=0.137\textwidth]{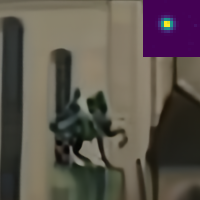}~
	    & \multicolumn{1}{c}{\hspace{-0.22cm}\makecell{\vspace{-2.55cm}  ~ \\\fbox{\shortstack[c]{\vspace{0.8cm} \\  No Ground-Truth\\ \hspace{0.41cm}(Real Image) \hspace{0.41cm} \\\vspace{0.65cm}}}}}  \\
	    
\multicolumn{1}{c|}{PSNR (dB)} & 20.76/20.33/- & 23.43/28.42/- & 22.41/17.99/- & 20.48/16.69/- & 25.74/31.23/- & \multicolumn{1}{|c}{26.11/33.79/-} \vspace{0.1cm}\\

\multicolumn{1}{c|}{\makecell{LR ($\times$4)}} & RCAN~\cite{zhang2018rcan} & DIP~\cite{ulyanov2018dip}  & \makecell{DIP~\cite{ulyanov2018dip}+Softmax \\+ USRNet~\cite{zhang2020usrnet}} & \makecell{Double-DIP~\cite{gandelsman2019doubledip,ren2020selfdeblur} \\+ USRNet~\cite{zhang2020usrnet}} & \makecell{\textbf{DIP-FKP} (ours) \\+ USRNet~\cite{zhang2020usrnet}}  & \multicolumn{1}{|c}{\makecell{GT + USRNet~\cite{zhang2020usrnet}}} \\
\end{tabular}
\vspace{-0.2cm}
\caption{Visual results of different methods on synthetic and real-world images for scale factor 4. Estimated/ground-truth kernels are shown on the top right of images. More visual results are provided in the supplementary.}
\label{fig:dipfkp_visualresults}
\end{figure*}

\subsection{Experiments on DIP-FKP}
\subsubsection{Quantitative Results}
\vspace{-0.1cm}
\paragraph{Comparison with state-of-the-arts.} Average PSNR and SSIM results of different methods are shown in Table~\ref{tab:dipfkp_psnr}. We compare the proposed DIP-FKP with bicubic interpolation, RCAN~\cite{zhang2018rcan}, DIP~\cite{ulyanov2018dip}, Double-DIP~\cite{gandelsman2019doubledip, ren2020selfdeblur} and the upper bound model (non-blind USRNet~\cite{zhang2020usrnet} given ground-truth kernels). Specifically, RCAN is one of the representative bicubic SR oriented models. When the kernel deviates from the predefined bicubic kernel, its performance deteriorates seriously. Double-DIP tries to remedy kernel mismatch by incorporating an untrained fully-connected network (FCN) as the kernel prior. However, its performance is unsatisfactory and worse than DIP that has no kernel prior. In contrast, DIP-FKP incorporates FKP as the kernel prior and improves the performance of DIP by significant margins across all datasets and scale factors. After applying the estimated kernels of DIP-FKP for non-blind SR, the performance could be further improved due to accurate kernel estimation.

\vspace{-0.3cm}
\paragraph{Comparison with other kernel priors.}
Table~\ref{tab:kp_compare_robustness_psnr} shows the results on kernel PSNR and image PSNR/SSIM (before and after non-blind SR) when using different kernel priors. ``DIP+Softmax'' uses a Softmax layer as the kernel prior, which is used to meet the non-negative and sum-to-one constraints on kernel. However, such a simple kernel prior gives rise to poor performance. Double-DIP adds two fully-connected layers before the Softmax layer, but its performance is still unsatisfactory. This actually accords with our analysis in Sec.~\ref{sec:original_doubledip} that an untrained network may not be a good kernel prior. For the special case of Gaussian kernel, it is also possible to use a parametric Gaussian generation model as a kernel prior (denoted as ``DIP+Parametric Prior''). As we can see, before non-blind SR, using FKP achieves similar image PSNR as using the parametric prior. This is because the quality of generated images is heavily dependent on DIP. However, FKP significantly outperforms the parametric prior in terms of kernel estimation, which leads to better image PSNR after non-blind SR. It is worth pointing out that, unlike the parametric prior, FKP can be used to model arbitrary kernel distributions as it is trained in an unsupervised way.

\begin{table}[!t]
\captionsetup{font=small}%
\scriptsize
\center
\begin{center}
\caption{Average PSNR/SSIM of using different kernel priors. Experiments are conducted on BSD100~\cite{BSD100} when scale factor is 2. Results on non-Gaussian kernel and image noise are also provided.}
\label{tab:kp_compare_robustness_psnr}
\begin{tabular}{|l|c|c|c|}
\hline
\multirow{3}{*}{Method} & \multirow{3}{*}{\makecell{Kernel \\PSNR}}  & \multicolumn{2}{c|}{Image PSNR/SSIM}\\
\cline{3-4}
& & \makecell{Before\\Non-Blind SR} & \makecell{After\\Non-Blind SR} 
\\
\hline
\hline
\multicolumn{4}{|c|}{$\times$2} \\
\hline
DIP~\cite{ulyanov2018dip} + Softmax & 32.67 & 23.62/0.5587 & 23.76/0.5783  \\
Double-DIP~\cite{gandelsman2019doubledip, ren2020selfdeblur}  & 39.98 &  23.31/0.5681 & 18.47/0.4441 \\
DIP~\cite{ulyanov2018dip} + Parametric Prior & 34.99 & \textbf{26.76/0.7091} & 28.00/0.7682 \\
DIP-FKP (ours) & \textbf{46.79} & {26.72/0.7089} & \textbf{28.61/0.8206}
\\
\hline  
\hline
\multicolumn{4}{|c|}{$\times$2, with Non-Gaussian Kernel} \\
\hline
DIP~\cite{ulyanov2018dip} + Softmax  & 32.84 & 23.69/0.5629 & 23.81/0.5877 \\
Double-DIP~\cite{gandelsman2019doubledip, ren2020selfdeblur}  & 39.36 & 23.29/0.5693 & 18.25/0.4364 \\
DIP~\cite{ulyanov2018dip} + Parametric Prior & 34.50 & 26.74/0.7096 &  27.77/0.7631\\
DIP-FKP (ours)  & \textbf{44.27} & \textbf{26.76/0.7097} & \textbf{27.88/0.8019} \\
\hline  
\hline
\multicolumn{4}{|c|}{$\times$2, with Image Noise of Level 10 (3.92\%)} \\
\hline
DIP~\cite{ulyanov2018dip} + Softmax  & 32.47 & 23.06/0.5314 & 23.67/0.5846 \\
Double-DIP~\cite{gandelsman2019doubledip, ren2020selfdeblur}  & 39.89 & 22.73/0.5322 & 21.95/0.6011\\
DIP~\cite{ulyanov2018dip} + Parametric Prior & 31.98 & 26.61/0.6939 & 27.11/0.7118\\
DIP-FKP (ours)  & \textbf{45.20} & \textbf{26.66/0.6946} & \textbf{27.67/0.7403} 
\\
\hline   
\end{tabular}
\end{center}
\end{table}

\subfigcapskip=-0.15cm
\begin{figure}[!t]
\captionsetup{font=small}
\begin{center}
\subfigure[\hspace{-0.5cm}]{\includegraphics[width=0.235\textwidth]{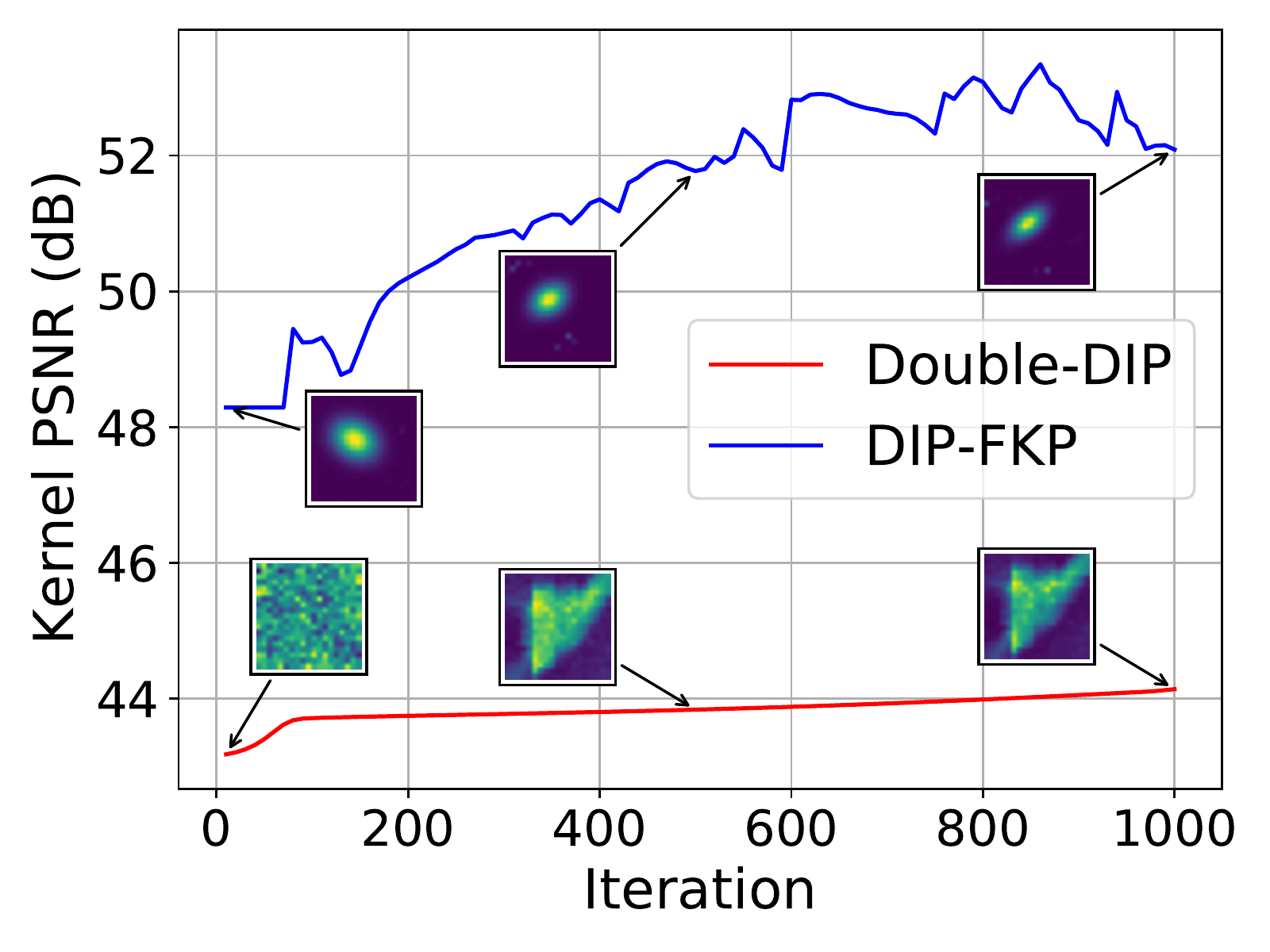}\label{fig:dipfkp_optimization}} 
\subfigure[\hspace{-0.5cm}]{\includegraphics[width=0.235\textwidth]{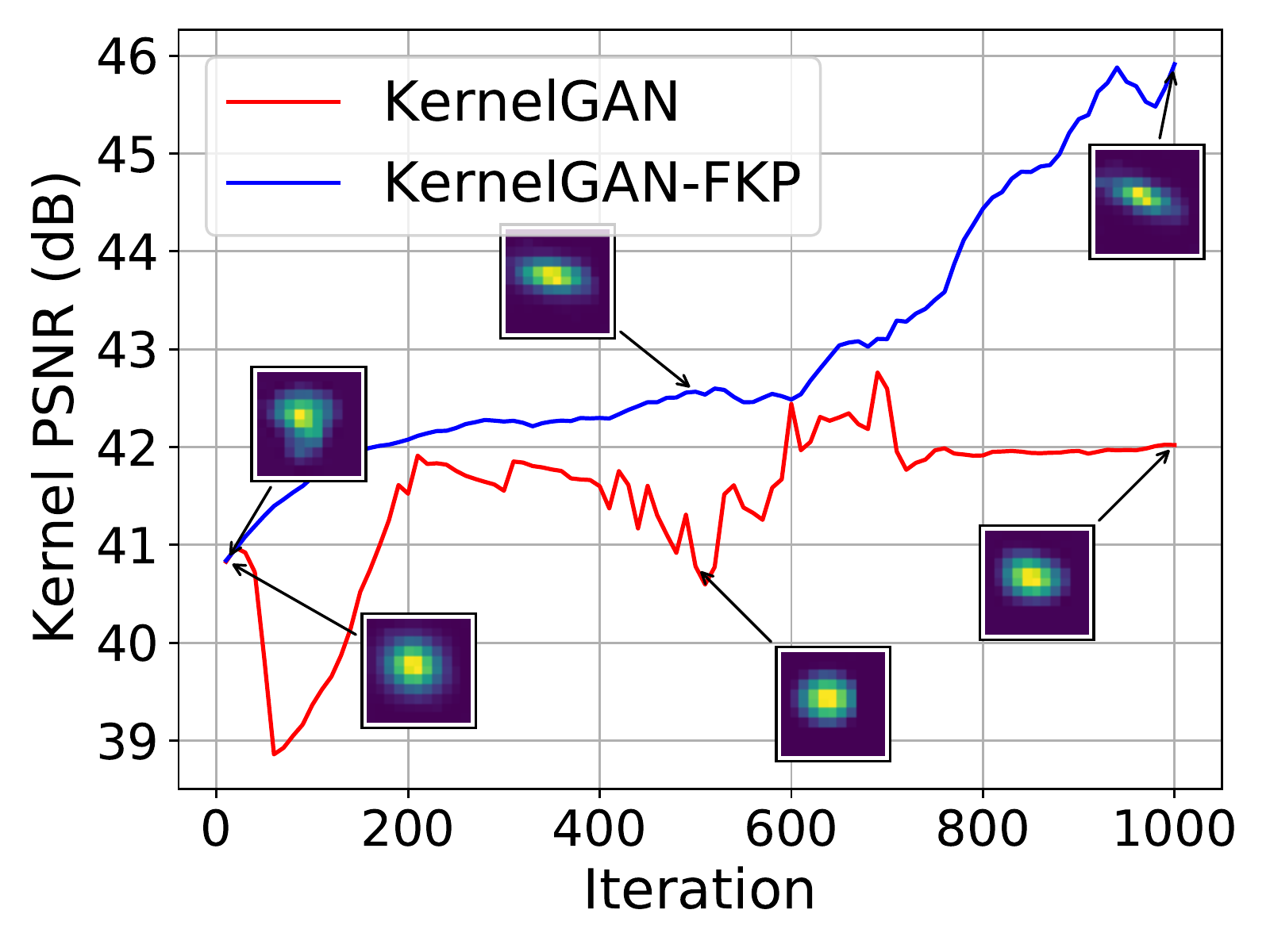}\label{fig:kernelganfkp_optimization}} 
\end{center}\vspace{-0.6cm}
\caption{The intermediate kernel results of Double-DIP, DIP-FKP, KernelGAN and KernelGAN-FKP during optimization. The testing images in (a) and (b) are ``\emph{156065}'' in BSD100~\cite{BSD100} and ``\emph{0812}'' in DIV2K~\cite{DIV2K}, respectively.}
\label{fig:dipfkp_kernelganfkp_optimization}
\vspace{-0.1cm}
\end{figure}
\subfigcapskip=0.1cm

\vspace{-0.3cm}
\paragraph{Robustness to non-Gaussian kernel and image noise.} 
We add noise to kernels and LR images to test the model robustness to non-Gaussian kernels and image noises, respectively. For non-Gaussian kernels, we apply uniform multiplicative noise (40\% of the maximum kernel pixel value, \ie, 0.4) on the kernel and then normalize it to meet the sum-to-one constraint. For image noise, we add noise of level 10 (3.92\% of the maximum image pixel value) to the image after blurring and downsampling. As one can see from Table~\ref{tab:kp_compare_robustness_psnr}, even under heavy kernel corruption, DIP-FKP still produces comparable results, showing good robustness to non-Gaussian kernels. When the image is corrupted by noise, DIP-FKP has a moderate performance drop, but it still outperforms its competitors by a large margin. In this case, we argue that the kernel estimation performance is mainly limited by DIP rather than the proposed FKP.

\vspace{-0.3cm}
\paragraph{Model parameter, runtime and memory usage.} The total number of parameters of kernel priors in DIP-FKP and Double-DIP are 143K and 641K, respectively. With a lightweight FKP, runtime and memory usage of DIP-FKP on a Tesla V100 GPU for generating a HR image of size $1,024\times 1,024$ are about 280 seconds and 10.6GB, while Double-DIP needs about 300 seconds and 11.2GB memory for the same setting.

\subsubsection{Visual Results}
\vspace{-0.1cm}
\paragraph{Comparison with state-of-the-arts.} 
The visual results of different methods on synthetic and real-world images are shown in Fig.~\ref{fig:dipfkp_visualresults}. As one can see, the results of RCAN tends to be blurry, while DIP tends to generate noise-like images. When different kernel priors are incorporated to DIP, both ``DIP+Softmax'' and Double-DIP fail to generate reasonable kernels, resulting in obvious artifacts such as over-smooth patterns and ringings. In contrast, DIP-FKP produces kernels that are very close to the ground-truths and generates the most visually-pleasant SR results for both synthetic and real-world images.

\vspace{-0.3cm}
\paragraph{Intermediate results during optimization.}
Fig.~\ref{fig:dipfkp_optimization} provides an example to show the intermediate kernel results of Double-DIP and DIP-FKP. It can be observed that Double-DIP is randomly initialized and only has a slight improvement during optimization. In contrast, with the incorporation of FKP, DIP-FKP has a good kernel initialization and the kernel is constrained to be in the learned kernel manifold, thereby converging better than Double-DIP.

\begin{table}[!tbp]
\captionsetup{font=small}
\scriptsize
\center
\begin{center}
\caption{Average PSNR/SSIM of different methods on DIV2K~\cite{DIV2K}. Note that KernelGAN is not applicable for scale factor 3. Results on small-image datasets are omitted due to weak patch recurrence.}
\label{tab:kernelganfkp_psnr}
\begin{tabular}{|l|c|c|}
\hline
Method  & \makecell{Kernel \\ PSNR} & \makecell{Non-blind \\ PSNR/SSIM}
\\
\hline
\hline
\multicolumn{3}{|c|}{$\times$2} \\
\hline
Bicubic Interpolation  & - & 26.97/0.7665\\
RCAN~\cite{zhang2018rcan}  & - & 26.99/0.7666\\ KernelGAN~\cite{ulyanov2018dip}+USRNet~\cite{zhang2020usrnet}  & 44.95 & 27.59/0.8162\\
KernelGAN-FKP + USRNet~\cite{zhang2020usrnet} (ours)  & \textbf{47.78} & \textbf{28.69/0.8567}\\
\hdashline
GT + USRNet~\cite{zhang2020usrnet} (upper bound)  & - & 34.59/0.9268 \\
\hline
\hline
\multicolumn{3}{|c|}{$\times$4} \\
\hline
Bicubic Interpolation & -  & 23.20/0.6329\\
RCAN~\cite{zhang2018rcan}  & - & 23.20/0.6310\\
KernelGAN~\cite{ulyanov2018dip}+USRNet~\cite{zhang2020usrnet} & 57.26  & 23.69/0.6539 \\
KernelGAN-FKP + USRNet~\cite{zhang2020usrnet} (ours) & \textbf{60.61}  & \textbf{25.46/0.7229}\\
\hdashline
GT + USRNet~\cite{zhang2020usrnet} (upper bound)  & - & 29.46/0.8069 \\
\hline
\hline
\multicolumn{3}{|c|}{$\times$2, with Non-Gaussian Kernel} \\
\hline
Bicubic Interpolation  & - & 26.96/0.7662 \\
RCAN~\cite{zhang2018rcan}  & - & 26.98/0.7663 \\ KernelGAN~\cite{ulyanov2018dip}+USRNet~\cite{zhang2020usrnet}  & 43.27 & 27.00/0.8030\\
KernelGAN-FKP + USRNet~\cite{zhang2020usrnet} (ours)  & \textbf{44.72} & \textbf{27.40/0.8334}\\
\hdashline
GT + USRNet~\cite{zhang2020usrnet} (upper bound)  & - & 34.59/0.9272   \\
\hline
\hline
\multicolumn{3}{|c|}{$\times$2, with Image Noise of Level 10 (3.92\%)} \\
\hline
Bicubic Interpolation  & - & 26.65/0.7258 \\
RCAN~\cite{zhang2018rcan}  & - & 26.22/0.6627\\ KernelGAN~\cite{ulyanov2018dip}+USRNet~\cite{zhang2020usrnet}  & 44.55 & 28.53/0.8281\\
KernelGAN-FKP + USRNet~\cite{zhang2020usrnet} (ours)  & \textbf{47.13} & \textbf{29.58/0.8303}\\
\hdashline
GT + USRNet~\cite{zhang2020usrnet} (upper bound)  & - & 31.27/0.8497 \\
\hline
\end{tabular}
\end{center}
\end{table}

\begin{figure}[!tbp]
\captionsetup{font=small}
\scriptsize
\hspace{-0.25cm}
\begin{tabular}{c@{\extracolsep{0em}}c@{\extracolsep{0em}}c@{\extracolsep{0em}}m{0em}m{0.27cm}}
        \includegraphics[width=0.137\textwidth]{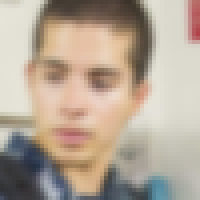}~
        &\includegraphics[width=0.137\textwidth]{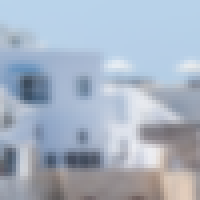}~
        & \hspace{-0.18cm}\includegraphics[width=0.137\textwidth]{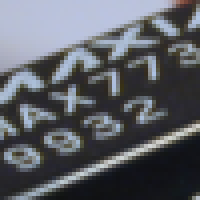} 
        & {\vspace{-2.3cm}\rotatebox{270}{\makecell{PSNR (dB)\vspace{-0.15cm}}}}
        & {\vspace{-2.3cm}\rotatebox{270}{\makecell{LR ($\times$4)}}}\\
        
        \cline{4-5}
        \includegraphics[width=0.137\textwidth]{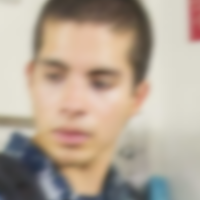}~
        &\includegraphics[width=0.137\textwidth]{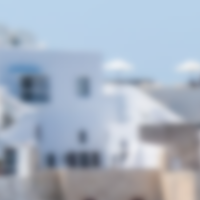}~
        & \hspace{-0.18cm}\includegraphics[width=0.137\textwidth]{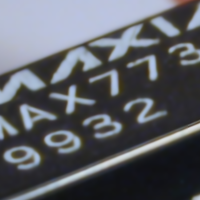} 
        & {\vspace{-2.3cm}\rotatebox{270}{\makecell{24.05/21.94/-\vspace{-0.15cm}}}}
        & {\vspace{-2.3cm}\rotatebox{270}{\makecell{RCAN~\cite{zhang2018rcan}}}}\\
        
        \includegraphics[width=0.137\textwidth]{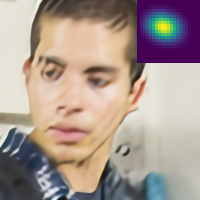}~
        &\includegraphics[width=0.137\textwidth]{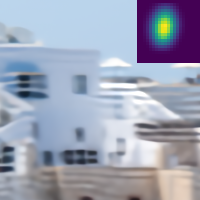}~
        & \hspace{-0.18cm}\includegraphics[width=0.137\textwidth]{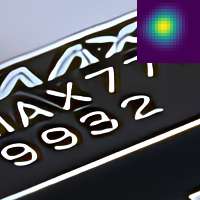} 
        & {\vspace{-2.3cm}\rotatebox{270}{\makecell{28.30/23.78/-\vspace{-0.15cm}}}}
        & {\vspace{-2.3cm}\rotatebox{270}{\makecell{KernelGAN~\cite{bell2019kernelgan}\\+ USRNet~\cite{zhang2020usrnet}\vspace{-0.15cm}}}}\\
        
        \includegraphics[width=0.137\textwidth]{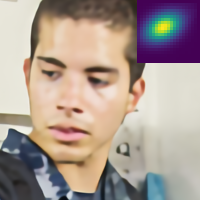}~
        &\includegraphics[width=0.137\textwidth]{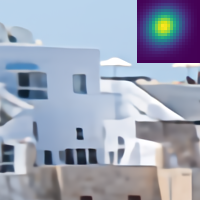}~
        & \hspace{-0.18cm}\includegraphics[width=0.137\textwidth]{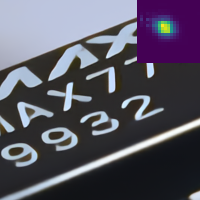} 
        & {\vspace{-2.3cm}\rotatebox{270}{\hspace{-0.2cm}\makecell{31.35/26.01/-\vspace{-0.15cm}}}}
        & {\vspace{-2.3cm}\rotatebox{270}{\hspace{-0.2cm}\makecell{\textbf{KernelGAN-FKP}(ours)\\+ USRNet~\cite{zhang2020usrnet}\vspace{-0.15cm}}}}\\
        
        \cline{4-5}
        \includegraphics[width=0.137\textwidth]{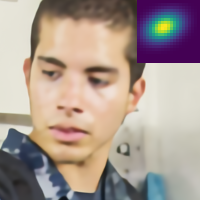}~
        &\includegraphics[width=0.137\textwidth]{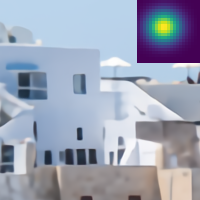}~
        & \multicolumn{1}{c}{\hspace{-0.22cm}\makecell{\vspace{-2.55cm} ~ \\
        \fbox{\shortstack[c]{\vspace{0.755cm} \\  No Ground-Truth\\ \hspace{0.38cm}(Real Image) \hspace{0.43cm} \\\vspace{0.65cm}}}
        }} 
        & {\vspace{-2.3cm}\rotatebox{270}{\makecell{32.18/26.81/-\vspace{-0.15cm}}}}
        & {\vspace{-2.3cm}\rotatebox{270}{GT + USRNet~\cite{zhang2020usrnet}}}\\
	
\end{tabular}
\caption{Visual results of different methods on synthetic and real-world images for scale factor 4. Estimated/ground-truth kernels are shown on the top right of images. More visual results are provided in the supplementary.}
\label{fig:kernelganfkp_visualresults}
\end{figure}

\subsection{Experiments on KernelGAN-FKP}
\subsubsection{Quantitative Results}
\vspace{-0.1cm}
\paragraph{Comparison with state-of-the-arts.} 
In Table~\ref{tab:kernelganfkp_psnr}, we compare the average kernel and image PSNR/SSIM of the proposed KernelGAN-FKP with bicubic interpolation, RCAN~\cite{zhang2018rcan}, KernelGAN~\cite{bell2019kernelgan} and the upper bound model (non-blind SR model USRNet~\cite{zhang2020usrnet} given ground-truth kernels). As one can see, RCAN has similar performance to naive bicubic interpolation due to kernel mismatch. KernelGAN is able to deal with different kernels and achieves better results than RCAN. Compared with KernelGAN, KernelGAN-FKP obtains further kernel PSNR gains of 2.83dB and 3.35dB for scale factors 2 and 4, respectively, which incur 1.1dB and 1.77dB image PSNR improvements after non-blind SR. The improvements can be attributed to the incorporation of FKP.

\vspace{-0.3cm}
\paragraph{Robustness to non-Gaussian kernel and image noise.}
Table~\ref{tab:kernelganfkp_psnr} also shows the results of different methods when given non-Gaussian kernels or noisy images. The experiment details are similar to DIP-FKP. As one can see, although all methods suffer from performance drops, KernelGAN-FKP still outperforms other methods by substantial margins. In particular, KernelGAN-FKP delivers comparable performance in handling image noise. The underlying reason might be that noise injection in the generator of GAN can help to circumvent over-fitting~\cite{feng2020noise}.

\vspace{-0.3cm}
\paragraph{Model parameter, runtime and memory usage.}
From KernelGAN to KernelGAN-FKP, the deep linear network is replaced by the proposed FKP, which reduces the total number of generator parameters from 151K to 143K. On a Tesla V100 GPU, runtime and memory usage of KernelGAN are about 93 seconds and 1.3GB, respectively, which are independent of images sizes. As for KernelGAN-FKP, it requires 90 seconds and 1.5GB memory.

\subsubsection{Visual Results}
\vspace{-0.1cm}
\paragraph{Comparison with state-of-the-arts.} Fig.~\ref{fig:kernelganfkp_visualresults} shows the visual comparison of different methods on synthetic and real-world images. It can be seen that RCAN tends to generate blurry images that are only slightly better than LR images. This is because the assumed bicubic kernel is sharper than the ground-truth kernels. Instead, KernelGAN tends to produce smoother kernels, leading to over-sharpen edges. In comparison, KernelGAN-FKP generates more accurate blur kernels and results in less artifacts for both synthetic and real-world images.

\vspace{-0.3cm}
\paragraph{Intermediate results during optimization.}
The intermediate kernel results of KernelGAN-FKP and KernelGAN are shown in Fig.~\ref{fig:kernelganfkp_optimization}. As one can see, KernelGAN-FKP converges more stably and better than KernelGAN, which oscillates during optimization. This suggests that the incorporation of FKP can increase the training stability and improve kernel estimation performance.

\subsection{DIP-FKP v.s. KernelGAN-FKP}
While DIP-FKP and KernelGAN-FKP outperform Double-DIP and KernelGAN, respectively, it is interesting to compare their differences. Since DIP-FKP jointly estimates the kernel and HR image, it requires more memory. Also, it generally generates better kernel estimations and has more stable convergence for small images. On the contrary, KernelGAN-FKP requires much less memory as it only needs to optimize the kernel, but it does not perform well for small images and large scale factors because it needs to re-downscale the LR image.

\section{Conclusion}
In this paper, we propose a flow-based kernel prior (FKP) for kernel distribution modeling and incorporate it into existing blind SR methods for better kernel and image estimation performance. FKP learns an invertible mapping between the complex kernel distribution and a tractable latent variable distribution based on normalization flow blocks. Its training is unsupervised and thus FKP is applicable for arbitrary kernel assumptions. When used as a kernel prior, FKP freezes its parameters and optimizes the latent variable in the network input space. Therefore, reasonable kernels are guaranteed for initialization and along optimization. FKP can be easily incorporated into existing kernel estimation models, such as Double-DIP and KernelGAN, by replacing their kernel modeling modules. Extensive experiments on synthetic LR images and real-world images demonstrate that FKP significantly improves the accuracy of kernel estimation and thus leads to state-of-the-art blind SR results.

\vspace{0.4cm}
\noindent\textbf{Acknowledgements}~~ This work was partially supported by the ETH Zurich Fund (OK), a Huawei Technologies Oy (Finland) project, the China Scholarship Council and a Microsoft Azure grant. Special thanks goes to Yijue Chen.

{\small
\bibliographystyle{ieee_fullname}
\bibliography{superresolution.bib}
}

\end{document}